\documentclass{article}

\PassOptionsToPackage{numbers, compress, sort}{natbib}
\usepackage[sort]{natbib}

\usepackage[preprint]{neurips_2024}

\usepackage{bbding}
\makeatletter
  \newcommand\figcaption{\def\@captype{figure}\caption}
  \newcommand\tabcaption{\def\@captype{table}\caption}
\makeatother

\usepackage{algorithm}
\usepackage{algorithmic}
\usepackage[utf8]{inputenc} 
\usepackage[T1]{fontenc}    
\usepackage{url}            
\usepackage{booktabs}       
\usepackage{amsfonts}       
\usepackage{nicefrac}       
\usepackage{microtype}      
\usepackage{array}

\newcolumntype{C}[1]{>{\centering\arraybackslash}m{#1}}
\usepackage{xcolor}         
\usepackage{color, colortbl}
\definecolor{citecolor}{HTML}{2980b9}
\definecolor{linkcolor}{HTML}{c0392b}
\definecolor{darkorange}{HTML}{FF8C00}
\definecolor{chocolate}{HTML}{D2691E}
\definecolor{darkgreen}{HTML}{006400}
\definecolor{darkblue}{HTML}{00008B}
\definecolor{mediumblue}{HTML}{0000CD}
\definecolor{dodgerblue}{HTML}{1E90FF}
\definecolor{royalblue}{HTML}{4169E1}
\definecolor{shadecolor}{RGB}{237,237,237}
\definecolor{backred}{RGB}{255, 190, 190}
\definecolor{backblue}{RGB}{210, 230, 250}

\definecolor{zrrgreen}{HTML}{008000}
\definecolor{zrrblue}{HTML}{4682B4}
\definecolor{zrrred}{HTML}{B22222}

\usepackage{amsmath}
\usepackage{amssymb}
\usepackage{graphicx}

\usepackage{multirow}
\usepackage{makecell}
\usepackage{caption}

\usepackage{adjustbox}
\usepackage{wrapfig}

\usepackage{float}
\usepackage{mmstyle}

\usepackage[hidelinks,breaklinks=true,colorlinks,bookmarks=false,citecolor=citecolor,linkcolor=linkcolor]{hyperref}


\usepackage{framed}
\usepackage{makecell}
\usepackage{listings}
\definecolor{lightgray}{rgb}{.9,.9,.9}
\definecolor{darkgray}{rgb}{.4,.4,.4}
\definecolor{purple}{rgb}{0.65, 0.12, 0.82}
\lstdefinelanguage{JavaScript}{
  keywords={break, case, catch, continue, debugger, default, delete, do, else, false, finally, for, function, if, in, instanceof, new, null, return, switch, this, throw, true, try, typeof, var, void, while, with},
  morecomment=[l]{//},
  morecomment=[s]{/*}{*/},
  morestring=[b]',
  morestring=[b]",
  ndkeywords={class, export, boolean, throw, implements, import, this},
  keywordstyle=\color{blue}\bfseries,
  ndkeywordstyle=\color{darkgray}\bfseries,
  identifierstyle=\color{black},
  commentstyle=\color{purple}\ttfamily,
  stringstyle=\color{red}\ttfamily,
  sensitive=true
}
\usepackage{xspace}
\usepackage[shortlabels]{enumitem}

\usepackage[most]{tcolorbox}
\tcbset{
  aibox/.style={
    width=474.18663pt,
    top=10pt,
    colback=white,
    colframe=black,
    colbacktitle=black,
    enhanced,
    center,
    attach boxed title to top left={yshift=-0.1in,xshift=0.15in},
    boxed title style={boxrule=0pt,colframe=white,},
  }
}
\newtcolorbox{AIbox}[2][]{aibox,title=#2,#1}

\usepackage{colortbl}
\usepackage{authblk}
\usepackage{xcolor}
\definecolor{orangeish}{HTML}{FFCC99}
\definecolor{blueish}{HTML}{87C2FF}

\usepackage{CJK}
\usepackage{caption}
\usepackage{graphicx}
\usepackage{multicol}
\usepackage{multirow}
\usepackage{booktabs}
\usepackage{amsmath}
\usepackage{upgreek}
\usepackage{footmisc}
\usepackage{tablefootnote}
\usepackage{footnote}
\usepackage{tabularx}
\usepackage{diagbox}
\usepackage{fancyvrb}
\usepackage{spverbatim}
\usepackage{nicematrix}
\usepackage{colortbl}
\usepackage{authblk}
\usepackage{xcolor}
\usepackage{amsmath}
\usepackage{amssymb}
\usepackage{mathtools}
\usepackage{amsthm}
\usepackage{multirow}
\usepackage{lscape}
\usepackage{listings}
\usepackage{subfigure}

\theoremstyle{plain}
\newtheorem{theorem}{Theorem}[section]

\newtheorem{lemma}[theorem]{Lemma}

\theoremstyle{definition}

\theoremstyle{remark}

\usepackage{stfloats}

\newcommand\blfootnote[1]{%
  \begingroup
  \renewcommand\thefootnote{}\footnote{#1}%
  \addtocounter{footnote}{-1}%
  \endgroup
}

\newcommand\methodname{OREAL}

\title{Exploring the Limit of Outcome Reward \\ for Learning Mathematical Reasoning}

\author{
    Chengqi Lyu$^{1*}$\quad Songyang Gao$^{1*}$\quad Yuzhe Gu$^{1,2*}$\quad Wenwei Zhang$^{1*\dag}$\quad Jianfei Gao$^{1}$ \\\vspace{-9pt}
    Kuikun Liu$^{1}$\quad Ziyi Wang$^{1}$\quad Shuaibin Li$^{1}$\quad Qian Zhao$^{1}$\quad Haian Huang$^{1}$\quad Weihan Cao$^{1}$\\
    Jiangning Liu$^{1}$ \quad Hongwei Liu$^{1}$\quad Junnan Liu$^{1}$\quad Songyang Zhang$^{1}$ \\
    Dahua Lin$^{1,3,4}$\quad Kai Chen$^{1\dag}$ \\\vspace{3pt}
    \small{$^1$Shanghai AI Laboratory} \quad \small{$^2$Shanghai Jiao Tong University}  \\
    \small{$^3$MMLab, The Chinese University of Hong Kong} \quad
    \small{$^4$HKGAI under InnoHK} \\
    \small\texttt{\{lvchengqi,gaosongyang,guyuzhe,zhangwenwei,chenkai\}@pjlab.org.cn}
}

\begin{document}
\blfootnote{$^*$ Equal contribution  $\dag$ Corresponding author}

\vspace{-0.6cm}
\maketitle

\begin{abstract}
Reasoning abilities, especially those for solving complex math problems, are crucial components of general intelligence.
Recent advances by proprietary companies, such as o-series models of OpenAI, have made remarkable progress on reasoning tasks. However, the complete technical details remain unrevealed, and the techniques that are believed certainly to be adopted are only reinforcement learning (RL) and the long chain of thoughts.
This paper proposes a new RL framework, termed \methodname{}, to pursue the performance limit that can be achieved through \textbf{O}utcome \textbf{RE}w\textbf{A}rd-based reinforcement \textbf{L}earning for mathematical reasoning tasks, where only binary outcome rewards are easily accessible.
We theoretically prove that behavior cloning on positive trajectories from best-of-N (BoN) sampling is sufficient to learn the KL-regularized optimal policy in binary feedback environments.
This formulation further implies that the rewards of negative samples should be further reshaped to ensure the gradient consistency between positive and negative samples.
To alleviate the long-existing difficulties brought by sparse rewards in RL, which are even exacerbated by the partial correctness of the long chain of thought for reasoning tasks, we further apply a token-level reward model to sample important tokens in reasoning trajectories for learning.
With \methodname{}, for the first time, a 7B model can obtain 94.0 pass@1 accuracy on MATH-500 through RL, being on par with 32B models. \methodname{}-32B also surpasses previous 32B models trained by distillation with 95.0 pass@1 accuracy on MATH-500.
Our investigation also indicates the importance of initial policy models and training queries for RL.
Code, models, and data will be released to benefit future research\footnote{https://github.com/InternLM/\methodname{}}.
\end{abstract}


\section{Introduction}
\label{sec: intro}

Solving complex problems with reasoning capability forms one of the cornerstones of human cognition - a cognitive ability that artificial general intelligence must ultimately master~\cite{xu2025towards, zhong2024evaluation}. Among various problem domains, the mathematical problem emerges as a particularly compelling experimental paradigm for AI research~\cite{liu2023mathematical, matzakos2023learning, ying2024internlm, shao2024deepseekmath}, owing to its relatively well-defined structure and availability of precise binary correctness feedback based on the verifiable final answers.

Recent advances in large language models (LLMs) have achieved remarkable progress in mathematical reasoning by the chain-of-thought technics~\cite{wei2022chain, wang2022self, kojima2022large}, 
in which the LLMs are elicited to produce a series of intermediate reasoning steps before providing the final answers to the problem.
However, as most of the capable models (\eg, the o-series models by OpenAI~\cite{openai2024learning}) are developed by proprietary companies, there is no clear pathway to develop state-of-the-art reasoning models. Some recent work shows that distillation~\cite{huang2024o1, Slow_Thinking_with_LLMs_2} is sufficient to obtain high performance given the accessibility to existing best or near best AI models, reinforcement learning (RL) is believed to be a more fundamental approach and has exhibited potential~\cite{deepseekr1} to advance beyond the intelligence boundary of current AI models, using the most capable open-source foundation models (DeepSeek-V3-base~\cite{liu2024deepseek}, \textit{inter alia}).

However, fundamental challenges of sparse reward in RL persist and are even exacerbated in mathematical reasoning tasks that mainly rely on the chain of thought technics with LLMs~\cite{wei2022chain}: the evaluation of intermediate reasoning steps is labor intensive~\cite{lightman2023let} and its accurate automation approach is still under-explored, thus, the only reliable reward is based on the outcome (correctness of final answer), which is inherently binary and sparse when faced with more than 2000 tokens in the long reasoning trajectories~\cite{deepseekr1, team2025kimi}. Existing approaches have attempted to estimate the advantages or values of reasoning steps by search~\cite{wang2024math, kazemnejad2024vineppo} or value function-based credit assignment~\cite{schulman2017proximal, cui2025process}, yet, their performance remains unsatisfactory in comparison with the distilled models~\cite{deepseekr1}.

This paper aims to conquer the above challenges and proposes a simple framework, termed \methodname{}, to push the limit of \textbf{O}utcome \textbf{RE}w\textbf{A}rd-based reinforcement \textbf{L}earning for mathematical reasoning tasks. \methodname{} is grounded in the unique characteristics of mathematical reasoning tasks that binary outcome feedback creates an environment where all positive trajectories are equally valid. We first establish that behavior cloning on BoN-sampled positive trajectories is sufficient to achieve KL-regularized optimality, which emerges from the analysis that the positive trajectory from BoN sampling converges to a distribution independent of the sample number. For learning on negative samples, \methodname{} reveals the necessity of reward shaping to maintain consistent gradient estimation between sampling and target distributions. Such a mechanism compensates for BoN's under-sampling of negative gradients, and enables difficulty-adaptive optimization over both successful and failed trajectories.

Another intrinsic property of mathematical reasoning tasks is the partial correctness in long reasoning chains, which further imposes the learning difficulty of sparse rewards when only a binary outcome reward is available at each iteration of RL training.
Thus, \methodname{} adopts a lightweight credit assignment scheme through a token-level reward model trained using outcome rewards. This mechanism automatically estimates step-wise importance weights by decomposing trajectory advantages, enabling focused learning of critical reasoning steps or errors.
The integration of these components yields a theoretically sound framework that effectively bridges the gap between sparse binary feedback and dense policy optimization requirements for mathematical reasoning tasks.

\begin{figure}[t]
    \centering
    \includegraphics[width=1.0\linewidth]{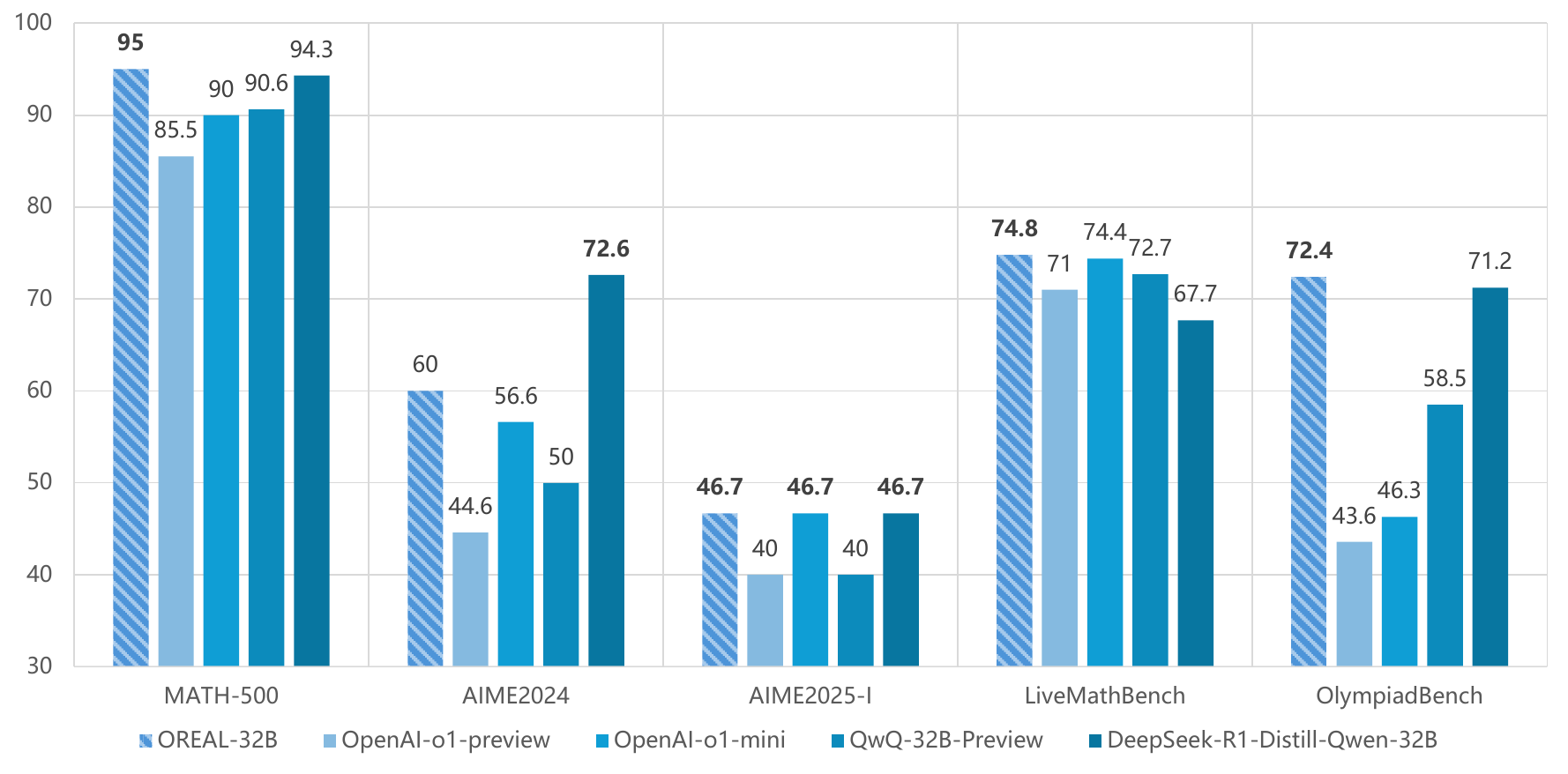}
    \caption{Overall performance between \methodname{}-32B and some competitive baselines.}
    \vspace{-16pt}
    \label{fig: main_fig}
\end{figure}

Extensive experimental results show that \methodname{} effectively improves the mathematical reasoning capability of LLMs.
At the 7B parameter scale, to the best of our knowledge, \methodname{}-7B is the first to obtain the pass@1 accuracy on MATH-500~\cite{hendrycks2021measuring} to 91.0 using RL instead of distillation, which even exceeds QwQ-32B-Preview~\cite{qwq-32b-preview} and o1-mini~\cite{openai2024learning}.
OREAL also improves DeepSeek-R1-Distilled-Qwen-7B from 92.8 to 94.0 pass@1 accuracy, being on par with the previous best 32B models.
For the 32B model, \methodname{}-32B outperforms all previous models (Figure~\ref{fig: main_fig}), both distilled and RL-based, obtaining new state-of-the-art results with 95.0 pass@1 accuracy on MATH-500.


\section{Methods}
\label{sec: methods}
To obtain a deeper understanding of the challenges when applying reinforcement learning (RL) for solving math word problems, we first analyze the formulation of RL and the intrinsic properties of underlying binary feedback environments (\S\ref{subsec:preliminary}), and establish a theoretical foundation for our optimization framework about how to learn from positive samples (\S\ref{subsec:learn_from_positive}) and failure trials (\S\ref{subsec:learn_from_negative}). To further conquer the learning ambiguity brought by outcome rewards to the partially correct long reasoning chains, we adopt a new strategy to estimate the importance of tokens for learning (\S\ref{subsec:tokenlevel_rm}).

\subsection{Preliminary}\label{subsec:preliminary}
When adopting a large language model (LLM) for mathematic reasoning, the input to the LLM policy is a textual math problem that prompts the LLM to output a multi-step reasoning trajectory consisting of multiple tokens as actions. During RL training, common practices~\cite{shao2024deepseekmath,yuan2023scaling} conduct sampling on the LLM to produce multiple reasoning trajectories, assign binary feedback (0/1 reward) based solely on the correctness of their final answer, and perform corresponding policy optimization using the sampled trajectories with reward.

\textbf{Policy Optimization.} Consider a Markov Decision Process (MDP) defined by the tuple \(( \mathcal{S}, \mathcal{A}, P, r, \gamma )\), where \(\mathcal{S}\) is a finite state space (\eg, contextual steps in mathematical reasoning), \(\mathcal{A}\) is the action space (\ie the token space of LLMs), \(P(s'|s, a)\) specifies the state transition dynamics, \(r: \mathcal{S} \times \mathcal{A} \rightarrow \mathbb{R}\) is the reward function, and \(\gamma \in [0,1)\) denotes the discount factor.  

In this section, we focus on KL-regularized policy optimization, which maximizes the expected cumulative returns while regularizing the policy \(\pi_\theta(\cdot|s)\) toward a reference policy \(\pi_0(\cdot|s)\). The objective function is formulated as:  

\begin{align}\label{eq: klpg}
J(\theta) \triangleq \mathbb{E}_{s \sim \rho_0, a \sim \pi_\theta(\cdot|s)}\left[ Q^{\pi_\theta}(s, a) \right] - \alpha \cdot \mathbb{E}_{s \sim \rho_0}\left[ D_{\text{KL}}\left(\pi_\theta(\cdot|s) \| \pi_0(\cdot|s)\right)\right]
\end{align}

with the state-action value function \(Q^{\pi}(s,a) = \mathbb{E}_{\pi}\left[\sum_{k=0}^\infty \gamma^k r(s_{t+k}, a_{t+k}) \mid s_t = s, a_t = a\right]\) under vanilla policy \(\pi\). This objective admits a closed-form solution for optimal policy \(\pi^*\):  
\begin{align}\label{eq: solution}
    \pi^*(a|s) = \frac{\pi_0(a|s) \exp\left( Q^{\pi}(s,a)/\alpha \right)}{Z(s)},
\end{align}

where \(Z(s) = \mathbb{E}_{a \sim \pi_0(\cdot|s)}\left[\exp\left( Q^{\pi}(s,a)/\alpha \right)\right]\) is the partition function that ensures normalization.  

\textbf{Best-of-N (BoN) Sampling.} As a common and efficient strategy to sample multiple reasoning trajectories from LLMs, Best-of-\(N\) sampling selects the trajectory with maximal reward among \(n\) independent rollouts from \(\pi_0\) to enhance policy performance. Formally, given candidate actions \(\{a^{(i)}\}_{i=1}^n \sim \pi_0(\cdot|s)\), the chosen action is \(a^* = \arg\max_{a^{(i)}} Q(s,a^{(i)})\). This strategy effectively leverages the exploration-exploitation trade-off through parallel sampling~\cite{gao2023scaling, go2023compositional}.  

\textbf{Binary Feedback under Outcome Supervision.}
Though a reasoning trajectory usually contains multiple reasoning steps with thousands of tokens, there lacks an efficient approach to automatically label the correctness of each token or reasoning step in math reasoning tasks.
Thus, a practical way is to parse the final answer from the reasoning trajectory~\cite{deepseekr1, lambert2024t}, evaluate its correctness based on rules or models, and then provide an outcome reward at the end of the trajectory as below:
\begin{align}\label{eq:orm}
    R(s_t) = \begin{cases} 
1 & \text{if \(t\) is the end step and the answer is correct} \\
0 & \text{otherwise},
\end{cases} 
\end{align}
which intrinsically treats the correct trajectory equally for learning. Moreover, the reward signal is severely sparse when compared to the thousands of tokens and does not provide any signal of progress or correctness for intermediate steps.
The resulting reward distribution of trajectories is also different from that of the dense reward function constructed through preference pairs in traditional RL for large language models~\cite{ouyang2022training}, which induces a more appropriate optimization framework for mathematical reasoning tasks, discussed in the next section.

\subsection{Learning from Positive Samples}\label{subsec:learn_from_positive}
Building upon the reward equivalence principle stated in Eq. \ref{eq:orm}, we first formalize a key probabilistic characteristic of BoN sampling:

\begin{lemma}\label{lemma:3.1}
Let $\pi(\theta, s)$ be a distribution over parameters $\theta$ and trajectory $s$, where each $s$ is associated with a binary reward $R(s) \in \{0, 1\}$. Define $p \triangleq \mathbb{E}_{s \sim \pi(\theta,\cdot)}[R(s) = 1] > 0$. Consider the BoN sampling:
$n = n_0 \to \infty$ and sample $\{s_1, s_2, \dots, s_n\}$ i.i.d. from $\pi_\theta$.
BoN selects $s^*$ uniformly from the subset with $R(s_i) = 1$. We have that,
The probability of selecting $s^*$ is converge to $\frac{\pi(\theta, s)}{p}$, which is independent of $n$.
\end{lemma}

The proof follows directly from the union law of BoN sampling ($\text{BoN}_{n+m} = \text{BoN}_2(\text{BoN}_m, \text{BoN}_n)$) and the trivial distinguishability of $0-1$ rewards. This result reveals that for problems with attainable positive responses, we are using a BoN generator with an arbitrary sampling budget to construct the positive training samples. 

To quantify the distributional divergence induced by BoN sampling, prior work~\cite{hilton2022measuring, scheurer2023training, coste2023reward} has analyzed the KL divergence between the BoN distribution $\pi_{\text{BoN}}$ and the original policy $\pi$. For continuous trajectory spaces $\mathcal{S}$, the BoN distribution admits the explicit form:
\begin{align}
\label{eq: pibon}
\pi_{\text{BoN}}(s) = n \cdot \left[P(s)\right]^{n-1} \cdot \pi(s),
\end{align}
where $P(s)$ denotes the cumulative distribution function (CDF) associated with $\pi(s)$. The corresponding KL divergence is given by
\[
\text{KL}(\pi_{\text{BoN}} \parallel \pi) = \log n - \frac{n-1}{n}.
\]

The KL divergence expression possesses two crucial properties, as $n \to \infty$, the $\text{KL}(\pi_{\text{BoN}}\parallel \pi)$ is strictly increasing in $n$ for $n \geq 1$, and converge to $\log n \to \infty$, covering the entire positive real axis.
This implies that for any prescribed KL divergence constraint $\epsilon > 0$, there exists a BoN parameterization for approximating the optimal policy, and can be simply sampled with the minimal $n(\epsilon)$ satisfying that
\[
n(\epsilon) = \arg\min_{n} \mathbb{E}_{s \sim \pi_{\text{BoN}}}[-R(s)].
\]

BoNBoN~\cite{gui2024bonbon} empirically shows that BoN sampling achieves the optimal win rate under fixed KL constraint by exhaustive search over the positive support. Therefore, the behavior cloning on BoN-selected positive samples directly learns the analytic solution to Eq. \ref{eq: klpg}. Intuitively, since every correct answer is preferred identically in the outcome-supervised sense, we only need to sample until we get a positive example, whose generating probability distribution will be the same as randomly picking from arbitrarily large numbers of samples.

Based on the theoretical understanding established, we formulate the first component of the learning objective in \methodname{} by incorporating KL-constrained max-likelihood-objective over positive examples obtained through sampling:

\[
\mathcal{L}_1(\theta) = \underbrace{\mathbb{E}_{s \sim \mathcal{D}^+} \left[ -\log \pi_\theta(s) \right]}_{\text{Positive example alignment}} + \beta \underbrace{\text{KL}(\pi_\theta \parallel \pi_{\text{old}})}_{\text{Policy constraint}},
\]
where $\mathcal{D}^+$ denotes the set of positive trajectories selected via BoN sampling from RL rollouts.

\subsection{Learning from Negative Samples}\label{subsec:learn_from_negative}
As established in Section \ref{subsec:learn_from_positive}, direct behavioral cloning on positive responses can effectively recover the policy distribution. BOND \cite{sessa2024bond} proposes estimating Jeffreys divergence \cite{jeffreys1946invariant} for the BoN strategy to train with both positive and negative samples, and demonstrates that signals from unsuccessful trajectories provide critical information about decision boundaries and failure modes.

In this section, we will discuss the relationship between the BoN (Best-of-N) distribution and the optimization objective defined in Eq. \ref{eq: klpg}, then elucidate the necessity of reward reshaping when training with negative samples.
Notably, while Eq. \ref{eq: pibon} shares structural similarities with Eq. \ref{eq: solution}, its application to mathematical reasoning tasks with binary feedback requires reformulation. Specifically, the transformed BoN distribution can be expressed as

\begin{align}\label{eq. pbon}
\pi_{\text{bon}}(s) = \pi(s) \left[ R(s) \cdot \frac{1 - \left(1 - p\right)^n}{p} + \left(1 - R(s)\right) \cdot \left(1 - p\right)^{n-1} \right],
\end{align}

which reveals fundamental differences between the BoN distribution and the original sampling distribution. Consider a scenario where two correct and two incorrect solutions are sampled, yielding an empirical accuracy of 50\%. However, the probability of selecting negative samples under Best-of-4 becomes $(0.5)^4 = 6.25\%$, significantly lower than the original distribution. This discrepancy necessitates reward shaping to maintain consistency between our optimization target and the expected return under the BoN distribution.

Building on BoN-RLB's \cite{chow2024inference} application of the log-likelihood trick for BoN-aware policy gradients, we analyze the reward shaping technic for negative samples to maintain gradient consistency with Section \ref{subsec:learn_from_positive}. With expectation return $p$ follow the definition in Lemma \ref{lemma:3.1}. The policy gradient under the BoN distribution can be derived as

\begin{equation}
\begin{aligned}
\nabla_\theta J_{\text{bon}} &= \mathbb{E}_{s \sim \pi_{\text{bon}}(\cdot)}\left[R(s) \nabla_\theta \log \pi_{\text{bon}}(s)\right] \\
&= \mathbb{E}_{s \sim \pi_{\text{bon}}(\cdot)}\left[R(s) \nabla_\theta\left( I_{D_+}(s)\log \pi(s)\frac{1 - (1-p)^n}{p} + I_{D_-}(s)\log \pi(s)(1-p)^{n-1} \right)\right],
\end{aligned}
\end{equation}

where $I_{D_+}(s)$ and $I_{D_-}(s)$ denote indicator functions for positive and negative sample sets respectively. Notably, these indicators are independent of policy parameters $\theta$. Given $\mathbb{E}_{s \sim \pi_{\text{bon}}}[I_{D_+}(s)] = 1-(1-p)^n$, we derive the gradient components as

\begin{align*}
    &\mathbb{E}_{s \sim \pi_{\text{bon}}}\left[\nabla_\theta \left( I_{D_+}(s) \log \pi(s)\frac{1 - (1-p)^n}{p}\right)\right] = n(1-p)^{n-1} \mathbb{E}_{s \sim \pi, s\in D_+}\left[\nabla_\theta \log \pi(s)\right].
\end{align*}
Similarly, we also have
\begin{align*}
\mathbb{E}_{s \sim \pi_{\text{bon}}}\left[\nabla_\theta \left( I_{D_-}(s) \log \pi(s)(1-p)^{n-1}\right)\right] = n(1-p)^n \mathbb{E}_{s \sim \pi, s\in D_-}\left[\nabla_\theta \log \pi(s)\right].
\end{align*}

This derivation reveals that when assigning unit reward ($R(s)=1$) to positive samples, gradient consistency requires reshaping negative sample rewards to $R^\star(s) \triangleq (1-p)R(s)$. Based on this reward shaping, we can construct policy optimization both on positive and negative samples for optimal policy.

To obtain the parameter $1-p$ which can be linked to the Monte-Carlo (MC) advantage estimation, we can simply estimate that probability by calculating the expected accuracy on the sample space by counting a small number of responses.
In this paper we apply a similar setting to RLOO \cite{fukunaga1989leave}, namely $R_{RLOO}(s) = R(s) - \frac{1}{N-1} \sum_{s^\star \neq s} R(s^\star)$ for unbiased mean reward
and train with policy gradient. The second part of our OREAL objective is then formulated as below:
\[
\mathcal{L}_{2}(\theta) = \mathbb{E}_{s \sim S_-} \left[  F(1-p) \cdot\log \frac{\pi_\theta(s)}{\pi_{old}(s)} \right]  + \beta \text{KL}(\pi_\theta \parallel \pi_{\text{old}}),
\]
where $p = \mathbb{P}_{\theta \sim \pi}[R(\theta) = 1]$, $S_-$ is the failed subset generated by policy model, and $F$ represents the preprocessing for advantage scores to serve as a generalized form, for example, $F(1-p) \triangleq \frac{r_i-mean(\{r_i...r_n\})}{std(\{r_i...r_n\})}$ in the recent GRPO \cite{shao2024deepseekmath} algorithm, where $mean(\{r_i...r_n\}) \to p$ when $n \to \infty$.

\subsection{Dealing with Long Reasoning Chains}\label{subsec:tokenlevel_rm}
In the previous discussion, we introduced the adaptation of binary reward training in response space. However, since the outcome supervision only provides feedback at the sequence level, this modeling essentially reduces to a contextual bandit without internal reward modeling within the MDP. A common counterexample is PPO, which utilizes a separate critic model to estimate the value function. However, such a solution appears to be expensive and complex, which has induced numerous explorations on how to stabilize the PPO training. 

Things become slightly different in mathematical reasoning, where the model can spontaneously revise omissions in intermediate steps to obtain the correct final answer. Therefore, outcome supervision is preferred, and the value function is more meant to be a simple credit assignment to determine how much the process step contributes to the outcome reward. With efficiency and performance trade-off considerations, we choose to use some low-cost alternatives for sequence-level reweighting.

Taking into account the deterministic dynamics in mathematical reasoning (\(s_{t+1} = f(s_t, a_t)\)), the state-action function $Q^\pi(s_{<t}, \pi(s_t))$ simplifies to the cumulative discounted reward of policy \(\pi\):  
\begin{align}\label{eq: vandq}
    Q^\pi(s_{<t}, \pi(s_t)) = V^\pi(s_{\leq t}) =  \sum_{k=0}^{T-t} \gamma^k r(s_{t+k} | s_{<t}).
\end{align}

Since intermediate rewards are not provided in mathematical reasoning tasks, we define an advantage function based solely on outcome feedback:
\begin{equation}
A(s_{\leq t}) = V^\pi(s_{\leq t+1}) - V^\pi(s_{\leq t}).
\end{equation}
This formulation treats \(A(s_{\leq t})\) as a token-wise credit assignment mechanism, estimating each token's contribution toward the final outcome.

For a pair of responses $y_1$ and $y_2$ to the same query, their initial values coincide $V_0^1 = V_0^2$. The win rate between them then satisfies:
\begin{equation}
\begin{aligned}\label{eq: winrate}
p(y_1>y_2)&=
\sigma(r(y_1) - r(y_2)) \\&= \sigma\left( \left(V_0^1 + \sum_{t=0}^T \gamma^t A_{y_1}^t\right) - \left(V_0^2 + \sum_{t=0}^T \gamma^t A_{y_2}^t\right) \right) \\
&= \sigma\left( \sum_{t=0}^T \gamma^t \left(A_{y_1}^t - A_{y_2}^t\right) \right).
\end{aligned}
\end{equation}

Equation \ref{eq: winrate} indicates that for any function family $\mathcal{A} = \{A(s_{\leq t})\}$ , a cumulative reward function through sequence aggregation can be constructed to model rewards:

\[
r^*(s) \triangleq \sum_{t=0}^{T} \gamma^t A(s_{\leq t}),
\]

which is trainable via preference pairs $\{(y_w, y_l)\}$ by fitting the outcome feedback. The learned $A(s_{\leq t})$ serves as a weighting function for credit assignment, which is used to reweight the original training loss, emphasizing critical reasoning steps or errors. An analogous implementations is $ r 2 Q^* $ \cite{rafailov2024r, xia2024inverse} by defining $A = \log \frac{\pi(y_i)}{\pi_{\text{ref}}(y_i)}$, PRIME \cite{cui2025process} then apply this formulation to improve performance of RLOO. In our work, following the practice from \cite{cobbe2021training}, we directly train a token-level reward function $w(s_{\leq t})$ satisfying
\[
\frac{1}{T}\sum_{t=0}^{T} w(s_{\leq t}) = r(s),
\]
without constraining KL-divergence to reference model in reward model training. These sequential rewards can serve as a proxy for the contribution of thinking steps to the result accuracy. Assuming a pair of prefix-consistent correct and incorrect samples, due to the causal inference nature of the token-level reward model, the preference optimization for these samples will only function on the steps that have different contents, which induces higher credits on the core reasoning step that affects the final result. We further discuss the training details of this model and analyze the visualization of its token-wise scoring effects later in Section \ref{subsec: rl} and Appendix \ref{sec: vis_token}.

In practice, we decompose the output weight $w(s)$ for positive and negative samples and clip on the positive axis to prevent reversing the direction of the optimized gradient, denoted as $\omega^+$ and $\omega^-$:

\begin{align}
\omega^+ = \max(2 \sigma(w) - 1, 0), \omega^- = \max(1 - 2\sigma(w), 0).
\end{align}

Giving input query $d$, the overall loss is as follows:

\begin{equation}
\begin{aligned}\label{eq: overall loss}
\mathcal{L}_{\text{total}}(d) \triangleq &\mathbb{E}_{s \sim S} \left[ \sum_{t=0}^T \left( -\omega_{s\leq t}^+\log \pi_{\theta}(s_{\leq t}|d)I_{D_+}(s) +  \eta 
 \ \omega_{s\leq t}^-\log \frac{\pi_\theta(s_{\leq t}|d)}{\pi_{old}(s_{\leq t}|d)}I_{D_-}(s) \right) \right] \\ &+ \beta \text{KL}(\pi_\theta(\cdot|d) \parallel \pi_{\text{old}}(\cdot|d)),
\end{aligned}
\end{equation}

where $\eta$ represents the balancing weights for positive and negative losses.

\section{Implementation}

\subsection{Policy Initialization}

We utilize Qwen2.5-7B and Qwen2.5-32B~\cite{yang2024qwen2} as the base model. 
Initially, we fine-tune the base models using long chain-of-thought data obtained through rejection sampling~\cite{yuan2023scaling}. 
This rejection sampling fine-tuned (RFT)~\cite{yuan2023scaling} models then serve as the initialization for the policy model in our RL framework.
We also explore to use of DeepSeek-R1-Distill-Qwen-7B~\cite{deepseekr1} as the initial policy model and perform \methodname{} on it and discuss the influence of different initial policy models in Section~\ref{subsec: policy model analysis}.
The training data for the RFT models consists of in-house datasets supported by OpenDataLab~\cite{opendatalab} and open-source datasets including Numina~\cite{li2024numinamath} and the training set of MATH~\cite{hendrycks2021measuring}.

\subsection{Reinforcement Learning}
\label{subsec: rl}

\noindent\textbf{Data Preparation.} During the on-policy RL process, we utilize questions from Numina, MATH training sets, and historical AMC/AIME (without AIME2024) competitions. For each question, we independently sample 16 trajectories from the RFT models. The correctness of each trajectory is then averaged to estimate the correctness rate of each query. To increase the difficulty of training queries, only questions with correctness rates between 0 and 0.8 are retained for further training.

\noindent\textbf{Outcome Reward Signal.} We employ the Qwen2.5-72B-Instruct~\cite{yang2024qwen2} as a generative verifier, in conjunction with a rule-based verifier, to evaluate the correctness of the model's outputs and provide binary rewards. This combination enhances the robustness of correctness assessment, mitigating issues related to the false negative of the rule-based verifier.

\noindent\textbf{Training Token-level Reward Model.} For the token-level reward model, we directly use the binary outcome rewards provided by the verifier and optimize using the cross-entropy loss:  

\begin{equation}
\mathcal{L}_{\text{CE}} = - \mathbb{E}_{(s, r) \sim \mathcal{D}} \left[ r \log p(s) + (1 - r) \log (1 - p(s)) \right],
\label{eq: ce}
\end{equation}

where $s$ represents the sampled trajectory, $ r \in \{0,1\}$  is the binary outcome reward from the verifier, and $p(s) = \sigma(\frac{1}{T}\sum_{t}^{T} w(s_{t}))$ denotes the predicted probability of correctness by the token-level reward model $w$.

To further analyze the behavior of the token-level reward model, we visualize its output distribution $w(s_{t})$ during the on-policy RL training process (see Appendix \ref{sec: vis_token}). In this training paradigm, $w(s_{t})$ assigns token-wise importance scores across the chain-of-thought reasoning process, capturing each token's contribution to the final correctness of the generated response. Consequently, this allows us to leverage $w(s_{t})$ for importance sampling during the optimization process, enabling a more principled selection of informative tokens. 

\noindent\textbf{Training Algorithm.} 
The loss function for the policy model follows the formulation described in Section \ref{sec: methods}. 
The complete RL training procedure is described in Algorithm \ref{alg:oreal}.

\begin{algorithm}[htbp]
\caption{The OREAL Reinforcement Learning Algorithm}
\begin{algorithmic}[1]
\STATE \textbf{Inputs:} Question set $\mathcal{D}$, policy model $\pi_\theta$, token-level reward model $w_\theta$, number of iterations $N$, batch size $B$, number of rollouts per question $K$.
\STATE Initialize policy $\pi_0$ and token-level reward model $w_0$ with $\pi_{\text{sft}}$.
\FOR{$i = 0, \dots , N$}
    \STATE Sample a batch of questions $\mathcal{D}_i \subseteq \mathcal{D}$ of size $B$.
    \STATE For $x \in \mathcal{D}_i$, generate $K$ policy samples: $Y = \{y_1, \dots, y_K\}$ where $y_k \sim \pi_i(x)$
    \STATE Obtain binary rewards $ \{r_1, \dots, r_K\}$ from verifier.
    \STATE Compute correctness rate: $p = \frac{1}{K} \sum_{k=1}^{K} r_k$ for reward shaping.
    \STATE Filter out questions with $0 < p < 1$ to avoid trivial cases.
    \STATE Select one correct $y^+$ and one incorrect sample $y^-$ for each question to avoid imbalance between positive and negative samples.
    \STATE Compute token-level importance sampling weights of each token with $w_i$.
    \STATE Use Eq \eqref{eq: ce} to update $w_i$.
    \STATE Update $\pi_i$ with Eq \eqref{eq: overall loss}
\ENDFOR
\STATE \textbf{Return:} The optimized policy model $\pi^*$.
\end{algorithmic}
\label{alg:oreal}
\end{algorithm}

\noindent\textbf{Hyperparameters.} 
The policy model is initialized from the RFT model. Similarly, the token-level reward model is also initialized with the same weights, but its output layer is replaced with a linear layer that produces a one-dimensional scalar. The weights of this layer are initialized to zero to ensure unbiased importance sampling weight at the start of training.  

During training iterations, each batch consists of 64 questions, with 16 rollouts per question. The max length of each rollout trajectory is set to 16384 tokens. Then the correctness of each response is averaged to calculate the pass rate, and questions with an overall pass rate of 0 or 1 are discarded. For the remaining trajectories, we retain only one correct response and one incorrect response per question, ensuring a balanced distribution of positive and negative samples for token-level reward model training. 

For optimization, the policy model is trained with a learning rate of $5e{-7}$, while the token-level reward model is trained with a learning rate of $2e{-6}$. The latter undergoes a 10-step warm-up phase before training begins. Both models employ a cosine annealing learning rate schedule, decaying to $1/5$ of the initial learning rate over time. We optimize both models using the AdamW optimizer. The total number of training steps is 80, with evaluation conducted every 10 steps. The KL coefficient $\beta$ is set to 0.01. We select the best-performing model determined by evaluation metrics.

\subsection{Skill-based Enhancement}
\label{subsec:skill-enhance}

During the RL training procedure, we observe that the model consistently struggles with certain types of questions, particularly those involving specific knowledge and skill areas, such as trigonometric constant transformations, probability statistics, series transformations, \etc. We believe this is caused by the insufficient learning of the base model on these concepts in the Pre-training or RFT stages.

To address this problem, we implement a skill-based enhancement approach, using the MATH dataset to reduce the high cost of skill annotation. Specifically, we annotate each question in the training set with its corresponding core skill. For questions that the model repeatedly fails to answer correctly during the RL phase, we perform data augmentation by including similar questions from the training set that share the same skill. These augmented questions are then added to the training data during the RFT stage to help the model better internalize these skills.

\section{Experiment}
\label{sec:experiment}

\subsection{Evaluation Setup}

\noindent\textbf{Baseline.}
We conduct evaluations against several baselines, including GPT-4o-0513~\cite{gpt4o}, Claude-Sonnet-3.5-1022~\cite{claude35sonnet}, OpenAI-o1-mini, OpenAI-o1-preview~\cite{openai2024learning}, Qwen2.5-Instrust-7B, Qwen2.5-Math-Instrust-7B, Qwen2.5-Instrust-32B~\cite{yang2024qwen2}, QwQ-32B-Preview~\cite{qwq-32b-preview}, DeepSeek-R1-Distill-Qwen-7B, DeepSeek-R1-Distill-Qwen-32B~\cite{deepseekr1}, SimpleRL~\cite{zeng2025simplerl}, PRIME~\cite{cui2025process}, rStarMath~\cite{guan2025rstar}.
For part of the baseline, we directly use the results from their report, which we mark with *.

\noindent\textbf{Benchmark.}
We use some well-established mathematical datasets for evaluation, including MATH-500~\cite{hendrycks2021measuring}, AIME2024~\cite{AIME2024}, AIME2025 (Part1)~\cite{AIME2024}, LiveMathBench~\cite{liu2024your}, and OlympiadBench~\cite{he2024olympiadbench}.

\noindent\textbf{Metrics.}
We use pass@1 as the metric for evaluation under the zero-shot chain-of-thought setting and use greedy decoding for each sample to assess correctness using OpenCompass~\cite{2023opencompass}.

\begin{table}[t]
\centering
\small
\begin{tabular}{l@{\hspace{0.1pt}}c@{\hspace{6pt}}c@{\hspace{6pt}}c@{\hspace{6pt}}c@{\hspace{6pt}}c}
\toprule
\textbf{Model} & \textbf{MATH-500} & \textbf{AIME2024} & \textbf{AIME2025-I} & \textbf{LiveMath} & \textbf{Olympiad} \\ \midrule
\multicolumn{6}{c}{API Models} \\ \midrule
GPT-4o-1120~\cite{gpt4o} & 72.8 & 16.7 & 13.3 & 44.8 & 33.7 \\
Claude-3.5-Sonnet-1022~\cite{claude35sonnet} & 78.3 & 13.3 & 3.3 & 46.7 & 35.4 \\
OpenAI-o1-preview~\cite{openai2024learning} & 85.5 & 44.6 & 40.0 & 71.0 & 43.6 \\
OpenAI-o1-mini~\cite{openai2024learning} & 90.0 & 56.6 & 46.7 & 74.4 & 46.3 \\ \midrule
\multicolumn{6}{c}{7B Models} \\ \midrule
Qwen2.5-Instrust-7B~\cite{yang2024qwen2} & 76.6 & 13.3 & 0.0 & 37.0 & 29.1 \\
Qwen2.5-Math-Instrust-7B~\cite{yang2024qwen2} & 81.8 & 20.0 & 13.3 & 44.1 & 31.1 \\
rStar-Math-7B~\cite{guan2025rstar}  & 78.4* & 26.7* & - & - & 47.1* \\
Qwen2.5-7B-SimpleRL~\cite{zeng2025simplerl}  & 82.4* & 26.7* & - & - & 37.6* \\
Eurus-2-7B-PRIME~\cite{cui2025process} & 79.2* & 26.7* & - & - & 42.1* \\
DeepSeek-R1-Distill-Qwen-7B~\cite{deepseekr1} & \underline{92.8}* & \textbf{55.5}* & \textbf{40.0} & \textbf{65.6} & \underline{64.1} \\
\textbf{\methodname{}-7B} & 91.0 & 33.3 & 33.3 & 62.6 & 59.9 \\ 
\textbf{\methodname{}-DSR1-Distill-Qwen-7B}  & \textbf{94.0} & \underline{50.0} & \textbf{40.0} & \textbf{65.6} & \textbf{66.1} \\  \midrule
\multicolumn{6}{c}{32B Models} \\ \midrule
Qwen2.5-Instrust-32B~\cite{yang2024qwen2} & 80.6 & 20.0 & 13.3 & 50.8 & 40.4 \\
QwQ-32B-Preview~\cite{qwq-32b-preview} & 90.6 & 50.0 & 40.0 & \underline{72.7} & 58.5 \\
DeepSeek-R1-Distill-Qwen-32B~\cite{deepseekr1} & 94.3* & \textbf{72.6}* & \textbf{46.7} & 67.7 & \underline{71.2} \\
\textbf{\methodname{}-32B} & \textbf{95.0} & \underline{60.0} & \textbf{46.7} & \textbf{74.8} & \textbf{72.4} \\ \bottomrule
\end{tabular}
\vspace{1em}
\caption{
Overall evaluation results for \methodname{} and each baseline.
``\methodname{}-DSR1-Distill-Qwen-7B'' denotes the DeepSeek-R1-Distill-Qwen7B trained by \methodname{}.
``AIME2025-I'', ``LiveMath'' and ``Olympiad'' represent ``AIME 2025 Part1'', ``LiveMathBench'', and ``OlympiadBench'', respectively.
For models at the parameter scale of 7B and 32B, we use Bold and Underlined to represent the best and second best performance, respectively.
For part of the baseline, we directly use the results from their report, marked with *.}
\vspace{-1em}
\label{tab: main_res}
\end{table}

\subsection{Overall Results}

Tabel~\ref{tab: main_res} shows the results of the comprehensive evaluation, highlighting the performance of our proposed models across different parameter scales. 
Notably, at the 7B scale, \methodname{}-7B achieves a remarkable pass@1 accuracy of 91.0 on the MATH-500 and 59.9 on OlympiadBench. 
To the best of our knowledge, this is the first time a model of this size has reached such a high level of accuracy using RL instead of distillation. 
This performance not only establishes a new milestone for RL-based methods but also surpasses significantly larger models, including QwQ-32B-Preview and OpenAI-o1-mini, demonstrating the effectiveness of our approach.
Furthermore, after applying \methodname{} on the previous best 7B model, DeepSeek-R1-Distill-Qwen-7B, the resulting model, OREAL-DSR1-Distill-Qwen-7B, obtains 94.0 and 66.1 pass@1 accuracy on MATH-500 and OlympiadBench, respectively, setting new records among all 7B models. This result verifies the effectiveness of \methodname{} even when faced with strong initial policies.

For 32B models, \methodname{}-32B achieves a groundbreaking pass@1 accuracy of 95.0 on MATH-500, 46.7 on AIME2025-I, 74.8 on LiveMathBench, and 72.4 on OlympiadBench, setting a new state-of-the-art among all previously reported models.
These results underscore the advantages of our methodology, including its scalability for training superior mathematical reasoning models across different model sizes.

Compared to the most competitive baseline, DeepSeek-R1-Distill-Qwen series, \methodname{}-32B demonstrates a clear advantage, whereas \methodname{}-7B lags slightly behind than the distilled 7B model, despite being trained on the same dataset as \methodname{}-32B. We attribute this discrepancy to the different affinities of the base models for the post-training data. Qwen-7B and Qwen-32B may exhibit varying degrees of knowledge gaps due to model sizes and pre-training settings. Our training data appears to better complement the existing knowledge of Qwen-32B, while it may be less effective in bridging gaps for Qwen-7B.

In addition, \methodname{}-DSR1-Distill-Qwen-7B improves the MATH-500 score from 92.8 to 94.0 and also achieves gains on LiveMathBench and OlympiadBench.
However, its performance on the AIME benchmark series is comparatively weaker. 
We observe the same disadvantages of \methodname{}-32B and \methodname{}-7B, whose AIME2024 scores are relatively lower than the best scores.
Since the overall performance verifies the effectiveness of the \methodname{} algorithm, we attribute the reason to the deficiency (\eg, in response quality, query difficulty, and quantity) of RFT data and RL training queries for obtaining high performance in the domain of AIME and leave it for the future work.

\subsection{Ablation Study}

\begin{figure}[t]
    \centering
    \begin{minipage}{0.45\linewidth}
        \centering
        \small
        \begin{tabular}{lc}
            \toprule
            \textbf{Setting} & \textbf{MATH-500}  \\ \midrule
            Initial Policy & 84.8  \\
            + REINFORCE (baseline) & 85.8  \\
            + Reward Shaping & 86.6  \\
            + Behavior Cloning & 87.6 \\
            + Importance Sampling & 89.0  \\ 
            + Skill-based Enhancement & 91.0 \\ \bottomrule
        \end{tabular}
        \vspace{1em}
        \captionof{table}{Ablation study for 7B models performance on MATH-500 with different reinforcement learning settings.}
        \label{tab: ablation}
    \end{minipage}
    \hfill
    \begin{minipage}{0.45\linewidth}
        \centering
        \includegraphics[width=\linewidth]{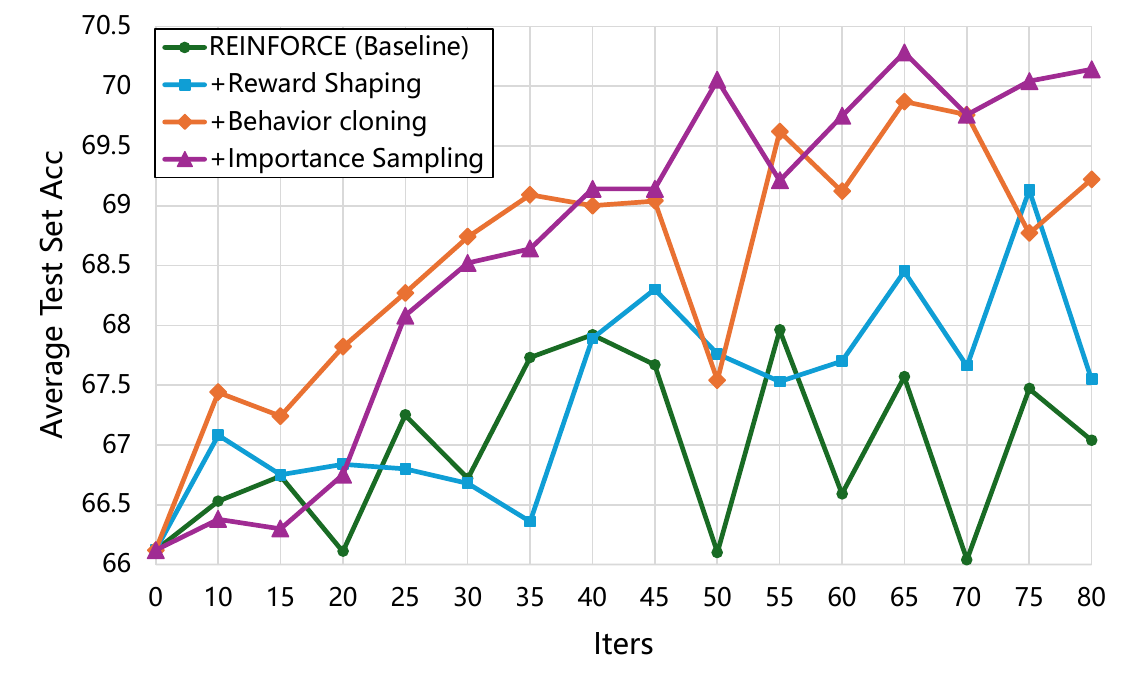}
        \vspace{-2em}
        \caption{Average test accuracy of 7B models across different training steps.}
        \label{fig: test acc}
    \end{minipage}
\end{figure}

To verify the effectiveness of each component described in Section~\ref{sec: methods}, we progressively add the proposed component based on the 7B model and compare the evaluation results on MATH-500, starting from REINFORCE~\cite{sutton1999policy} as baseline.

As shown in Tabel~\ref{tab: ablation}, we add each component step by step, where ``Reward Shaping'' represents $L_2$ introduced in Section~\ref{subsec:learn_from_negative}, ``Behavior Cloning'' represents $L_1$ introduced in Section~\ref{subsec:learn_from_positive}, ``Importance Shaping'' represents $L_{total}$ introduced in Section~\ref{subsec:tokenlevel_rm}.
The gradual addition of the modules steadily increases the Pass@1 scores of the 7B model on MATH-500, proving the effectiveness of our method.
Ultimately, the policy model is raised from an initial score of 84.8 to 91.0.

We also report average pass@1 accuracy across all benchmarks during the training process with different RL settings. 
As shown in Figure~\ref{fig: test acc}, the REINFORCE training process is unstable, which can be mitigated by ``Reward Shaping''.
``Behavioral Cloning'' for positive samples can speed up convergence and show better performance early in training.
Although the performance growth of ``Importance Sampling'' is relatively slow in the early stage of training, it ultimately obtains the best results.

\subsection{Analysis of Initial Policy Models}
\label{subsec: policy model analysis}

\begin{table}[t]
\centering
\small
\begin{tabular}{l@{\hspace{0.1pt}}c@{\hspace{6pt}}c@{\hspace{6pt}}c@{\hspace{6pt}}c@{\hspace{6pt}}c}
\toprule
\textbf{Model} & \textbf{MATH-500} & \textbf{AIME2024} & \textbf{AIME2025-I} & \textbf{LiveMath} & \textbf{Olympiad} \\ \midrule
OREAL-7B-SFT-wo-enhance & 84.8 & 26.7 & 26.7 & 55.0 & 55.1 \\
\methodname{}-7B-wo-enhance & 89.0 & 36.7 & 40.0 & 60.1 & 58.1 \\ 
\midrule
OREAL-7B-SFT & 86.4 & 26.7 & 26.7 & 54.2 & 56.0 \\
\methodname{}-7B & 91.0 & 33.3 & 33.3 & 62.6 & 59.9 \\ 
\midrule
DeepSeek-R1-Distill-Qwen-7B~\cite{deepseekr1} & 92.8* & 55.5* & 40.0 & 65.6 & 64.1 \\
\methodname{}-DSR1-Distill-Qwen-7B  & 94.0 & 50.0 & 40.0 & 65.6 & 66.1 \\ 
\midrule
OREAL-32B-SFT & 92.6 & 43.3 & 46.7 & 71.9 & 68.7 \\
OREAL-32B & 95.0 & 60.0 & 46.7 & 74.8 & 72.4 \\

\bottomrule
\end{tabular}
\vspace{1em}
\caption{Evaluation for the performance of \methodname{} on different initial policy models. Here, ``-SFT'' and ``DeepSeek-R1-Distill-Qwen7B'' denote the initial policy model. ``wo-enhance'' means the model which do not perform the skill-based enhancement during the SFT stage.}
\vspace{-1em}
\label{tab: base_model}
\end{table}

We further analyze \methodname{} by adopting it to several different initial policy models, as shown in Table~\ref{tab: base_model}. \methodname{} consistently improves the performance of each initial policy model, including our own trained model and the strong distilled model~\cite{deepseekr1}, on MATH-500, LiveMathBench, and OlympiadBench, except slight fluctuations on AIME2024 and AIME2025 part1 when the performance of initial policy models are already high (\eg, DeepSeek-R1-Distill-Qwen-7B), which demonstrates the generality of \methodname{}.

After adding skill-based enhancement data (introduced in Section~\ref{subsec:skill-enhance}), there is a significant rise in MATH-500 scores for the initial policy model (row 1 and row 3) and the corresponding RL-trained model (row 2 and row 4). 
Since our enhancement is performed primarily for the MATH-500, this verifies the effectiveness of the skill-based enhancement approach. In addition, the performance of the model after RL is strongly correlated with the capabilities of the initial policy model itself. 
The stronger the initial policy model, the higher the performance that RL can deliver, indicating the importance of policy initialization.


\section{Related Work}

\noindent\textbf{Stimulate Reasoning using Chain of Thought.}
In mathematical reasoning tasks, Chain of Thought (CoT)~\cite{wei2022chain} is recognized as a crucial technique to enhance the reasoning ability of large language models (LLMs), which can be implemented through few-shot examples~\cite{wei2022chain} or zero-shot prompt engineering~\cite{kojima2022large}. 
Self-consistency (SC)~\cite{wang2022self} is further proposed to generate and voting through multiple CoTs. 
In addition to simple CoTs, various search methods have been explored that simultaneously consider multiple potential CoTs, such as Tree-of-Thought (ToT)~\cite{yao2024tree} and Graph-of-Thought (GoT)~\cite{besta2024graph}, which extend the idea to tree or graph structure, offering more flexibility in developing CoTs and backtracking.
However, these methods mainly stimulate the reasoning capability of LLMs by prompts without parameter updates, these inference-time techniques do not fundamentally improve the underlying ability of LLMs.

\noindent\textbf{Reasoning Enhancement by Supervised Fine-tuning.}
To let the LLMs essentially acquire the reasoning abilities, many studies~\cite {ying2024internlm, yu2023metamath, liu2024augmenting, li2023query, liu2023goat, yue2023mammoth} have explored synthesizing high-quality data to conduct supervised fine-tuning (SFT) on LLMs. 
But this method heavily relies on high-quality training data and a existing high-performing model~\cite{lightman2023let}. 
As a result, many existing works~\cite{huang2024o1, Slow_Thinking_with_LLMs_2} have turned to distilling knowledge from powerful, large-scale models to synthesize data, yielding good results. 
However, distilled-based methods receive the limitations of the teacher model.
One criticism of SFT is its limited generalization capability~\cite{chu2025sft}. 
Some studies argue that SFT merely transforms the model into a knowledge retriever, rather than an actual reasoner~\cite{kambhampati2024can}.

\noindent\textbf{Reinforcement Learning for LLM.}
Compared to SFT, reinforcement learning (RL) offers better generalization and is therefore considered a more fundamental training aproach~\cite{chu2025sft}. 
Previous attempts applying RL for LLMs mainly target aligning the LLM to human preferences~\cite{ouyang2022training}.
Later, some works~\cite{ying2024internlm, shao2024deepseekmath, luo2023wizardmath, lightman2023let} has attempted to use it to enhance the model's reasoning and obtained promising results.
Recently, the advent of the o1 family of models~\cite{openai2024learning} and a series of o1-like works~\cite{deepseekr1, zeng2025simplerl, cui2025process, guan2025rstar} make the importance of large-scale RL for inference became more apparent. 
Currently, the mainstream approach to RL involves using outcome reward signals~\cite{deepseekr1, zeng2025simplerl, kazemnejad2024vineppo} and there are different views in the community on how to use that reward signal.
ReST$^{EM}$~\cite{singh2023beyond} and RFT~\cite{yuan2023scaling} simply select the positive samples based on the binary signal and only use them for behavior cloning.
GRPO~\cite{shao2024deepseekmath}, RLOO~\cite{fukunaga1989leave, ahmadian2024back}, REINFORCE~\cite{sutton1999policy}, use both positive and negative samples for policy updating, but facing the challenges of sparse reward in long sequence.
PPO~\cite{schulman2017proximal} makes the preference modeling on sequence-level.
Different from them, to explore the limit of outcome reward, \methodname{} presents a unified framework, grounded in the unique characteristics of mathematical reasoning tasks.

\section{Conclusion and Future Work}

This paper aims to explore the limit of \textbf{O}utcome \textbf{RE}w\textbf{A}rd-based reinforcement \textbf{L}earning for mathematical reasoning tasks, and proposes a unified policy optimization framework, termed \methodname{}, grounded in three key insights: 1) Behavior cloning on positive trajectories from Best-of-n (BoN) sampling is both necessary and sufficient for optimal policy learning under binary feedback;
2) Accordingly, a reward-shaping mechanism should be further introduced to transform the reward function derived from the optimal policy;
3) An efficient token-level credit assignment scheme can be achieved through trajectory advantage decomposition without relying on additional value networks.
Together, these components form a theoretically grounded, general, and scalable approach for mathematical reasoning tasks.
With \methodname{}, we are the first to improve the performance of a 7B model on the MATH-500 accuracy to 91 using the RL method instead of distillation, which even surpasses OpenAI-o1-mini and QwQ-32B-Preview. Even when taking the previous best 7B model, DeepSeek-R1-Distill-Qwen-7B, as initial policy, \methodname{} can improve it to 94 pass@1 accuracy on MATH-500, being even on par with the previous best 32B models.
\methodname{}-32B also obtains new state-of-the-art results among the 32B model on MATH-500, LiveMathBench, and OlympiadBench.

Along with the experimental observations presented in this paper, we also find two factors that are crucial for the success of scalable RL for mathematical reasoning tasks, which become the primary focus of our future work.
First, the initial policy model should be as free of knowledge deficiencies as possible, as this serves as the foundation for further improvement during the RL stage. A strong starting point ensures that RL can effectively and efficiently incentivize the underlying capability of LLMs obtained through pre-training or supervised fine-tuning. Towards this goal, it is a practical way to conduct distillation or data synthesis with DeepSeek-R1 or DeepSeek-V3, which is not explored in this work as it is orthogonal to our investigation.
Second, the quality of the data used in the RL phase must be diverse and sufficient in terms of difficulty, quantity, and scope. A well-balanced dataset enables the model to reach its full potential by exposing it to a broad range of challenges and learning opportunities. Thus, we believe it is still valuable to make efforts in the pre-training and post-training data construction process.


{
\bibliographystyle{unsrt}
\bibliography{neurips_2024}
}

\clearpage
\appendix
\setcounter{table}{0} 
\setcounter{figure}{0}
\setcounter{equation}{0}
\renewcommand{\thetable}{A\arabic{table}}
\renewcommand\thefigure{A\arabic{figure}} 
\renewcommand\theequation{A\arabic{equation}}

\section{Token Level Reward Model Score Visualization}
\label{sec: vis_token}

\begin{figure}[htbp]
    \centering

    \begin{subfigure}
        \centering
        \includegraphics[width=0.99\textwidth]{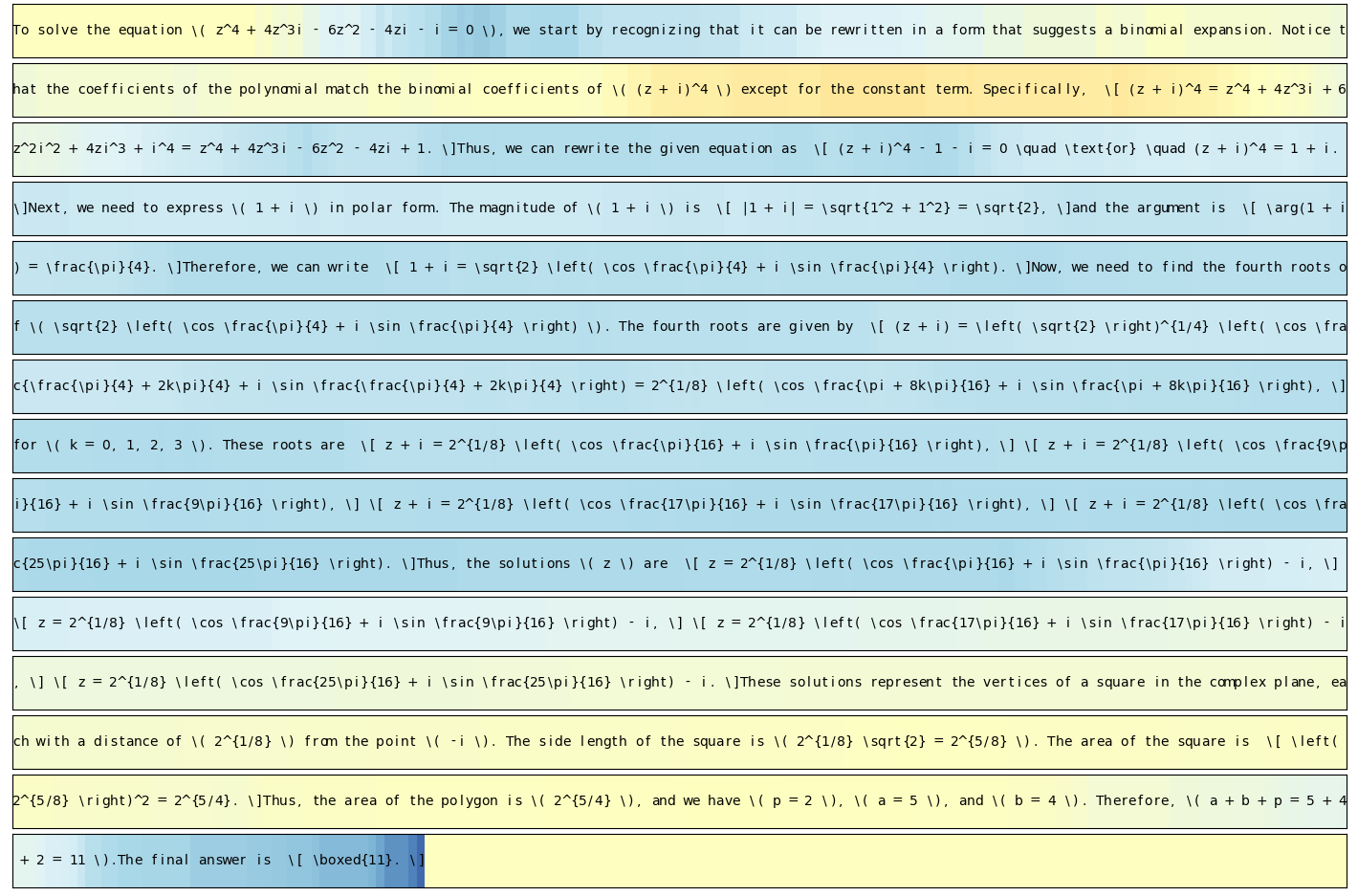} 
        \caption{Token-level reward model score visualization for a correct response.}
        \label{fig:token_score_correct}
    \end{subfigure}
    \hfill

    \begin{subfigure}
        \centering
        \includegraphics[width=0.99\textwidth]{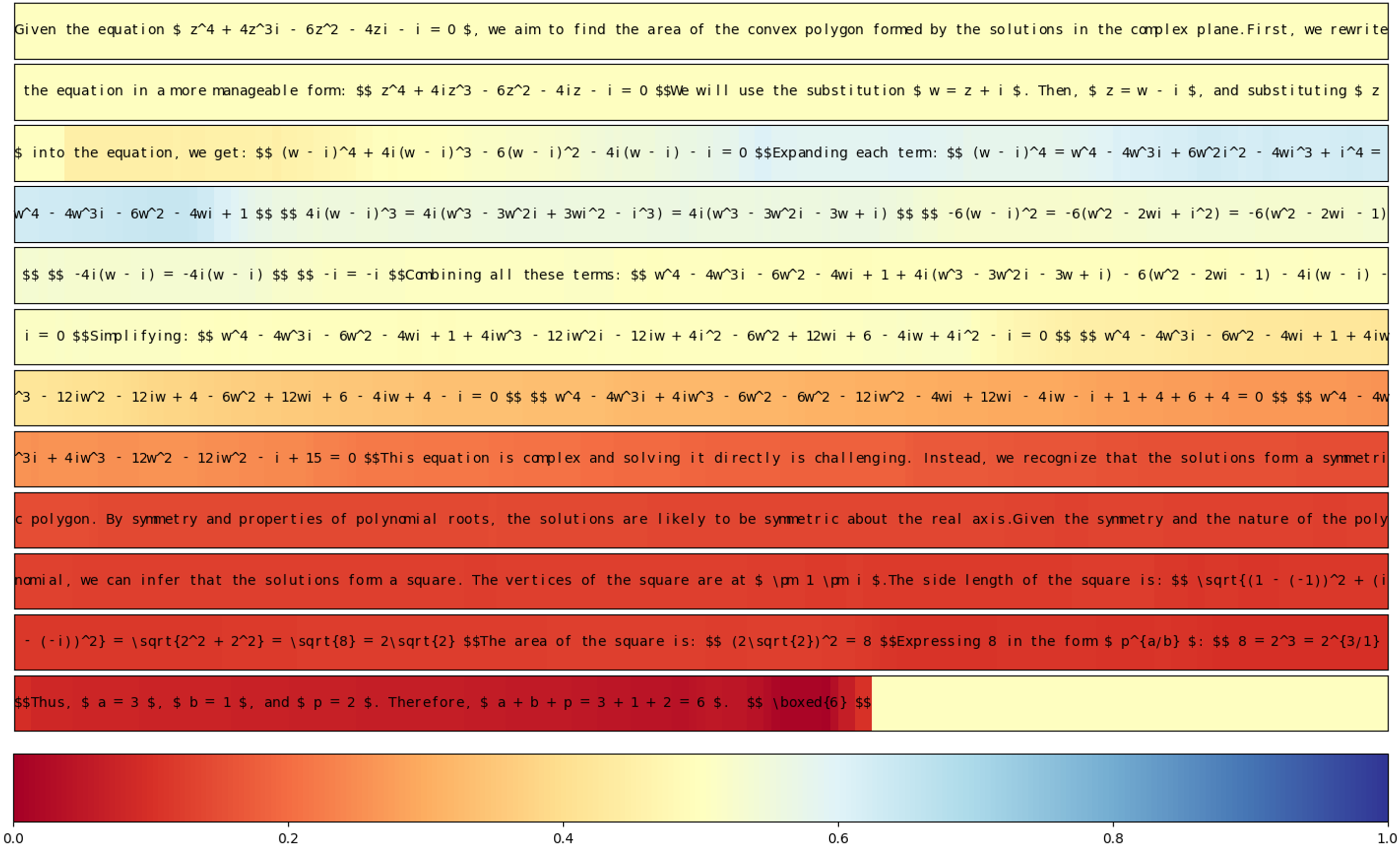}
        \caption{Token-level reward model score visualization for an incorrect response.}
        \label{fig:token_score_incorrect}
    \end{subfigure}

\end{figure}

Figure~\ref{fig:token_score_correct} and~\ref{fig:token_score_incorrect} show the token-level reward model scores across responses. The values are normalized to [0, 1]. Cooler colors indicate higher reward scores, while warmer colors denote lower scores.
For correct responses, the overall REWARDS are high, especially at the end, although there are a few lower sections in the middle. For incorrect responses, the distribution of rewards is reversed, and the closer to the end the lower the rewards.
This indicates that not all tokens contribute to the response equally and it is important to assign token-level credits to the sequences.

\section{Prompt}

Figure~\ref{fig: verifier prompt} is the system prompt of the verifier model, which is used during RL training to provide the binary outcome reward for a response.
Figure~\ref{fig: sys prompt} is the system prompt we use for fine-tuning and RL training as well as the evaluation.

\begin{figure*}[h] 
\begin{AIbox}{}
{\bf Verifier Prompt:} \\
{
You are a helpful assistant who evaluates the correctness and quality of models' outputs.
\\

Please as a grading expert, judge whether the final answers given by the candidates below are consistent with the standard answers, that is, whether the candidates answered correctly. 
\\
    
Here are some evaluation criteria:
\\

1. Please refer to the given standard answer. You don't need to re-generate the answer to the question because the standard answer has been given. You only need to judge whether the candidate's answer is consistent with the standard answer according to the form of the question. Don't try to answer the original question. You can assume that the standard answer is definitely correct.

2. Because the candidate's answer may be different from the standard answer in the form of expression, before making a judgment, please understand the question and the standard answer first, and then judge whether the candidate's answer is correct, but be careful not to try to answer the original question.

3. Some answers may contain multiple items, such as multiple-choice questions, multiple-select questions, fill-in-the-blank questions, etc. As long as the answer is the same as the standard answer, it is enough. For multiple-select questions and multiple-blank fill-in-the-blank questions, the candidate needs to answer all the corresponding options or blanks correctly to be considered correct.

4. Some answers may be expressed in different ways, such as some answers may be a mathematical expression, some answers may be a textual description, as long as the meaning expressed is the same. And some formulas are expressed in different ways, but they are equivalent and correct.

5. If the prediction is given with $\backslash$boxed\{\}, please ignore the $\backslash$boxed\{\} and only judge whether the candidate's answer is consistent with the standard answer.
\\

Please judge whether the following answers are consistent with the standard answer based on the above criteria. Grade the predicted answer of this new question as one of:

A: CORRECT 

B: INCORRECT

Just return the letters "A" or "B", with no text around it.
\\

Here is your task. Simply reply with either CORRECT, INCORRECT. Don't apologize or correct yourself if there was a mistake; we are just trying to grade the answer.

<Original Question Begin>: 

{ORIGINAL QUESTION}

<Original Question End>
\\

<Gold Target Begin>: 

{GOLD ANSWER}

<Gold Target End>
\\

<Predicted Answer Begin>: 

{ANSWER}

<Predicted End>
\\

Judging the correctness of candidates' answers:

}
\end{AIbox} 
\caption{Prompts for the model-based generative verifier.}
\label{fig: verifier prompt}
\end{figure*}

\begin{figure*}[h] 
\begin{AIbox}{}
{\bf System Prompt:} \\
{
You are an expert mathematician with extensive experience in mathematical competitions. You approach problems through systematic thinking and rigorous reasoning. When solving problems, follow these thought processes:
\\

\#\# Deep Understanding

Take time to fully comprehend the problem before attempting a solution. Consider:

- What is the real question being asked?

- What are the given conditions and what do they tell us?

- Are there any special restrictions or assumptions?

- Which information is crucial and which is supplementary?
\\

\#\# Multi-angle Analysis

Before solving, conduct through analysis:

- What mathematical concepts and properties are involved?

- Can you recall similar classic problems or solution methods?

- Would diagrams or tables help visualize the problem?

- Are there special cases that need separate consideration?
\\

\#\# Systematic Thinking

Plan your solution path:

- Propose multiple possible approaches

- Analyze the feasibility and merits of each method

- Choose the most appropriate method and explain why

- Break complex problems into smaller, manageable steps
\\

\#\# Rigorous Proof

During the solution process:

- Provide solid justification for each step

- Include detailed proofs for key conclusions

- Pay attention to logical connections

- Be vigilant about potential oversights
\\

\#\# Repeated Verification

After completing your solution:

- Verify your results satisfy all conditions

- Check for overlooked special cases

- Consider if the solution can be optimized or simplified

- Review your reasoning process
\\

Remember:

1. Take time to think thoroughly rather than rushing to an answer

2. Rigorously prove each key conclusion

3. Keep an open mind and try different approaches

4. Summarize valuable problem-solving methods

5. Maintain healthy skepticism and verify multiple times
\\

Your response should reflect deep mathematical understanding and precise logical thinking, making your solution path and reasoning clear to others.
When you're ready, present your complete solution with:

- Clear problem understanding

- Detailed solution process

- Key insights

- Thorough verification
\\

Focus on clear, logical progression of ideas and thorough explanation of your mathematical reasoning. Provide answers in the same language as the user asking the question, repeat the final answer using a '$\backslash$boxed\{\}' without any units, you have [[8192]] tokens to complete the answer.

}
\end{AIbox} 
\caption{System prompts for long CoT reasoning.}
\label{fig: sys prompt}
\end{figure*}

\clearpage

\end{document}